# A COMPARATIVE EVALUATION OF MACHINE LEARNING ALGORITHMS FOR THE PREDICTION OF R/C BUILDINGS' SEISMIC DAMAGE

Konstantinos Demertzis[1,2], Konstantinos Kostinakis[3], Konstantinos Morfidis[4], Lazaros Iliadis[2]

[1]School of Science & Technology, Informatics Studies, Hellenic Open University, Greece
e-mail: kdemertz@fmenr.duth.gr

[2]School of Engineering, Department of Civil Engineering, Faculty of Mathematics Programming and General Courses, Democritus University of Thrace, Kimmeria, Xanthi, Greece
e-mail: liliadis@civil.duth.gr

[3]Assistant Professor, Department of Civil Engineering, Aristotle University of Thessaloniki
Aristotle University campus, 54124, Thessaloniki, Greece
e-mail: kkostina@civil.auth.gr

[4]Assistant Researcher, Earthquake Planning and Protection Organization (EPPO-ITSAK)
Terma Dasylliou, 55535, Thessaloniki, Greece
e-mail: konmorf@gmail.com

**Corresponding author:**
*K. Kostinakis*
*Pasalidi 19, Kato Toumpa*
*54453 Thessaloniki, Greece*
*tel. +30 6945349502*
*email: kkostina@civil.auth.gr*

**Abstract.** Seismic assessment of buildings and determination of their structural damage is at the forefront of modern scientific research. Since now, several researchers have proposed a number of procedures, in an attempt to estimate the damage response of the buildings subjected to strong ground motions, without conducting time-consuming analyses. These procedures, e.g. construction of fragility curves, usually utilize methods based on the application of statistical theory. In the last decades, the increase of the computers' power has led to the development of modern soft computing methods based on the adoption of Machine Learning algorithms. The present paper attempts an extensive comparative evaluation of the capability of various Machine Learning methods to adequately predict the seismic response of R/C buildings. The training dataset is created by means of Nonlinear Time History Analyses of 90 3D R/C buildings with three different masonry infills' distributions, which are subjected to 65 earthquakes. The seismic damage is expressed in terms of the Maximum Interstory Drift Ratio. A large-scale comparison study is utilized by the most efficient Machine Learning algorithms. The experimentation shows that the LightGBM approach produces training stability, high overall performance and a remarkable coefficient of determination to estimate the ability to predict the buildings' damage response. Due to the extremely urgent issue, civil protection mechanisms need to incorporate in their technological systems scientific methodologies and appropriate technical or modeling tools such as the proposed one, which can offer valuable assistance in making optimal decisions.

# 1. Introduction

One of the most important, but also challenging, scientific issues in the field of earthquake engineering is the estimation of the structural response of buildings subjected to earthquake ground motions. Since now, numerous research studies have dealt with the above issue and proposed a vast variety of different methods aiming at the seismic assessment of structures. Many of these methods focus on the rapid determination of the earthquake damage response and on the seismic vulnerability assessment of large number of buildings without performing computationally hard analyses, in an attempt to overcome the difficulties resulting from the time-consuming conduction of demanding nonlinear analysis methods (e.g [1-7]), These procedures usually utilize methods based on the application of statistics theory. In the last decades, the increase of the computers' power has led to the development of modern statistical methods based on the adoption of Machine Learning (ML) algorithms. The up to date research on these methods revealed that they can provide a fast, reliable, and computationally easy way for screening of vulnerable structures and that they can be used as an efficient alternative to the conduction of demanding numerical simulations (e.g. [8-17]). The achievement of this goal is made through the creation of a relationship mapping that emulates the structure's behavior.

ML is one of the most important scientific field of the new era that includes those algorithmic methods that can be learned from data. It combines ideas from the sciences of statistics and probabilities to make accurate future predictions, while mathematical optimization techniques are used to improve the performance of a system. There are four distinct categories of ML with independent characteristics of learning: a) the information-based learning methodologies that employ concepts from information theory to build models, b) the similarity-based learning methods that build models based on comparing features of known and unknown objects or measure similarity between past and forthcoming occurrences, c) the probability-based learning techniques that build models based on measuring how likely it is that some event will occur and, finally, d) the error-based learning that builds models based on minimizing the total error through a set of training instances. On the other hand, based on how to use the data, there are three main categories of ML algorithms: a) Supervised Learning in which the training process of the algorithm is based on samples of labeled data, b) Unsupervised Learning which is the ability of the algorithm to detect patterns in unknown data and, finally, c) Reinforcement Learning that employs algorithms for discovering the environment based on rewarded actions.

Several research studies have proved that the ML methods, mainly Artificial Neural Networks (ANNs), can effectively assess the seismic response of complex structures. A comprehensive literature review of the most commonly used and newly developed ML techniques for the assessment of the buildings' damage has been made by Harirchian et. al [18], by Xie et. al [19] and by Sun et. al [20]. A brief review of some of the most important research works is given below. Molas and Yamazaki [21] were among the first researchers who studied the ability of ANNs to adequately predict the seismic damage of wooden structures. At almost the same time, Stephens and VanLuchene [22] used trained ANNs in order to estimate the damage level of R/C structures expressed by means of Park and Ang damage index. Rafiq et. al [23] investigated various types of ANNs (Multi-layer Perceptron, Radial Basis Networks and normalized Radial Basis Networks), aiming at utilizing them to solve engineering problems. In another significant research study conducted by Latour and Omenzetter [24] the ability of the ANNs to reliably estimate the earthquake-induced damage of planar R/C frames was investigated by using nonlinear time history analyses' results. A similar investigation was carried out by Arslan [25], who studied the impact of certain structural parameters on the damage level of regular R/C buildings under seismic ground motions. To this end, a dataset, created through the application of nonlinear pushover analyses, was used to train the ANNs. Rofooei et al. [26] utilized data from nonlinear dynamic analyses of 2D moment resisting R/C frames in order to investigate the influence of structural and seismic features on the ANNs' performance. The correlation between the interstory drift ratios and the plastic hinge rotation of R/C shear walls was studied by Vafei et al. [27], who used ANNs trained by results taken from nonlinear modal pushover analyses. Kia and Sensoy [28] investigated the impact of certain seismic parameters on the ability of ANNs to assess the seismic damage level of R/C concrete frames based on nonlinear time history analyses of a 2D moment resisting R/C frame. Kostinakis and Morfidis conducted a series of research studies [29-32] in an attempt to estimate the reliability of ANNs as regards the estimation of the seismic response of R/C buildings. In their studies, they examined also the number and the combination of the input parameters through which an optimum prediction for the damage state of R/C buildings can be achieved, the influence of the parameters which are used for the configuration of the networks' training on the efficiency of their predictions, as well as the impact of the presence of masonry infills on the results. Burton et al. [33] adopted ML methods in an attempt to estimate aftershock collapse vulnerability of buildings utilizing mainshock intensity, seismic response and certain damage indicators. More recently, Zhang et al. [34] proposed a ML framework for the assessment of the post-earthquake structural safety of a 4-story R/C special moment frame building. In another research conducted by the same research team [35] several ML methods were utilized in order to adequately estimate the residual structural capacity of damaged tall buildings. In another paper [36], a novel framework for earthquake vulnerability assessment of buildings via Rapid Visual Screening is proposed using type-2 fuzzy. Nguyen et al. [37] adopted ANNs and Extreme Gradient Boosting methods for the prediction of planar steel moment frames' seismic response. In particular, the researchers used a comprehensive dataset for the training and testing of the ML models, created by nonlinear dynamic analyses of 36 steel moment frames with different structural characteristics subjected to a large number of ground motions. In a most recent study, Li et al. [38] proposed a method that combines the interstory drift spectrum and a deep learning method to estimate the maximum interstory drift ratio of buildings.

The results of the most research studies established the ability of ML techniques in the successful prediction of the seismic damage. However, all of the abovementioned researchers adopted one or only a few ML methods for their study, namely, no study has made an attempt to utilize a large number of ML methods, in order to comparatively evaluate their efficiency in assessing the damage response with adequate reliability. The present paper aims at an extensive comparative evaluation of a large number of Machine Learning algorithms for the reliable prediction of 3D R/C buildings' seismic response. In order to accomplish this aim, a large training dataset consisting of 30 R/C buildings with different structural parameters (the number of stories, the structural eccentricity and the ratio of base shear received by R/C walls (if they exist) along the two orthogonal horizontal axes) was selected. The buildings were designed on the basis of provisions of EC2 [39] and EC8 [40]. For each one of these buildings three different configurations regarding their masonry infill walls were assumed (without masonry infills, with masonry infills in all stories and with masonry infills in all stories except for the ground story), leading to three different data subsets consisting of 30 buildings each. The selected buildings were analyzed for 65 appropriately chosen real earthquake records using Nonlinear Time History Analyses (NTHA). As inputs in the process of Machine Learning methods both seismic and structural parameters widely used in the literature were chosen. The well-documented Maximum Interstory Drift Ratio (MIDR) was selected as the damage index for the R/C.

## 2. Formulation of the problem in terms compatible to Machine Learning methods

### 2.1 Overview of the procedure
In this section the procedure adopted in order to formulate the problem in terms compatible to ML methods is presented. The procedure consists of the following steps:
- Generation of the training data set, which includes selection of a large number of representative R/C buildings, design and modeling of the inelastic properties of the them and selection of an adequate number of seismic motions.
- Selection of the problem's input (structural and seismic) parameters.
- Conduction of NTHA, according to which the buildings are analysed for the selected earthquake records and their seismic response is determined. Consequently, processing of the analyses' results in order to compute the values of an appropriate seismic damage index (in the present study the MIDR index), which is selected as the output parameter (target) of the ML procedures.

### 2.2 Training data set
In order to fulfill the purposes of the present research study, a large training data set consisting of buildings with a variety of structural characteristics was considered. An attempt was made to select structures that are representative of the buildings designed and built with the aid of modern seismic codes and according to the common construction practice in European countries with regions of high seismicity. More specifically, a set of 30 R/C buildings was selected (see [30]). The buildings' structural system consists of members in two perpendicular directions (denoted as axes x and y). Moreover, they are rectangular in plan and regular in elevation and in plan according to the criteria set by EC8 [40]. The buildings possess different characteristics concerning the stories' number $n_{st}$ (stories' height: 3.2m), the value of structural eccentricity $e_o$ (i.e. the distance between the mass center and the stiffness center of stories) and the ratio of the base shear received by the walls along two horizontal orthogonal directions (axes x and y): $n_{vx}$ and $n_{vy}$. The values of these structural parameters for the selected buildings are given in Table 1. More details about the selected buildings can be found in [30].

**Table 1** The values of structural parameters of the selected R/C buildings

| No. | $n_{st}$ | $L_x$(m) | $L_y$(m) | $e_o$(m) | $n_{vx}$ (%) | $n_{vy}$ (%) | No. | $n_{st}$ | $L_x$(m) | $L_y$(m) | $e_o$(m) | $n_{vx}$ (%) | $n_{vy}$ (%) |
|---|---|---|---|---|---|---|---|---|---|---|---|---|---|
| 1 | 3 | 13.5 | 10.0 | 0.0 | 0.0 | 0 | 16 | 3 | 13.0 | 9.0 | 0.98 | 0.0 | 0.0 |
| 2 | 5 | 20.0 | 14.0 | 0.0 | 0.0 | 0.0 | 17 | 5 | 17.5 | 10.0 | 2.58 | 0.0 | 0.0 |
| 3 | 7 | 20.0 | 14.0 | 0.0 | 0.0 | 0.0 | 18 | 7 | 17.5 | 10.0 | 2.39 | 0.0 | 0.0 |
| 4 | 3 | 15.0 | 10.0 | 0.0 | 73.0 | 76.0 | 19 | 3 | 13.5 | 9.0 | 4.65 | 52.0 | 46.0 |
| 5 | 5 | 19.0 | 16.2 | 0.0 | 77.0 | 80.0 | 20 | 5 | 16.0 | 14.5 | 4.19 | 43.0 | 42.0 |
| 6 | 7 | 19.0 | 16.2 | 0.0 | 57.0 | 64.0 | 21 | 7 | 16.0 | 14.5 | 3.79 | 37.0 | 36.0 |
| 7 | 3 | 15.0 | 15.0 | 0.0 | 41.0 | 41.0 | 22 | 3 | 13.5 | 9.0 | 2.23 | 47.0 | 0.0 |
| 8 | 5 | 21.2 | 18.7 | 0.0 | 46.0 | 50.0 | 23 | 5 | 16.0 | 14.5 | 2.65 | 38.0 | 0.0 |
| 9 | 7 | 21.2 | 18.7 | 0.0 | 43.0 | 46.0 | 24 | 7 | 16.0 | 14.5 | 2.49 | 35.0 | 0.0 |
| 10 | 3 | 17.0 | 12.5 | 0.0 | 43.0 | 0.0 | 25 | 3 | 14.5 | 9.0 | 3.53 | 64.0 | 0.0 |
| 11 | 5 | 20.2 | 15.2 | 0.0 | 41.0 | 0.0 | 26 | 5 | 14.0 | 16.0 | 3.01 | 0.0 | 69.0 |
| 12 | 7 | 20.2 | 15.2 | 0.0 | 38.0 | 0.0 | 27 | 7 | 14.0 | 16.0 | 3.01 | 0.0 | 65.0 |
| 13 | 3 | 15.0 | 10.0 | 0.0 | 77.0 | 0.0 | 28 | 3 | 13.5 | 10.0 | 6.73 | 64.0 | 58.0 |
| 14 | 5 | 20.2 | 15.2 | 0.0 | 68.0 | 0.0 | 29 | 5 | 16.5 | 16.5 | 6.29 | 65.0 | 72.0 |
| 15 | 7 | 20.2 | 15.2 | 0.0 | 51.0 | 0.0 | 30 | 7 | 16.5 | 16.5 | 5.96 | 59.0 | 67.0 |

In the above table, $L_x$ and $L_y$ are the dimensions of the rectangular shaped plans of the selected buildings and $e_0=(e_{0x}^2+e_{0y}^2)^{1/2}$, where $e_{0x}$, $e_{0y}$ are the structural eccentricities along axes x and y respectively.

In order to investigate the impact of the masonry infills on the seismic response and damage of the buildings, for each one of the 30 structures three different assumptions about the distribution of the masonry infills were considered, leading to three different training subsets: (a) subset denoted as ROW_FORM_BARE consisting of the 30 buildings without masonry infills (bare structures), (b) subset denoted as ROW_FORM_FULL-MASONRY consisting of the 30 buildings with masonry infills uniformly distributed along the height (infilled structures) and (c) subset denoted as ROW_FORM_PILOTIS consisting of the 30 buildings with the first story bare and the upper stories infilled (structures with pilotis). Consequently, the total number of structures investigated herein is 30 different structural systems x 3 different distributions of masonry infills = 90. The three abovementioned subsets of the buildings, as a result of their different masonry infills' configurations, were trained separately by the same ML methods, in order to draw conclusions about the possible differences in the predictive ability of the ML techniques, resulting from the influence of the infill walls on the seismic response of them.

The 30 selected bare buildings were modeled, analyzed and designed according to the provisions of EC2 [39] and EC8 [40]. For the buildings' elastic modelling all recommendations of EC8 were followed (diaphragmatic behavior of the slabs, rigid zones in the joint regions of beams/columns and beams/walls, values of flexural and shear stiffness corresponding to cracked R/C elements). The buildings were classified as Medium Ductility Class (MDC) structures. The analyses and design was done with the aid of the modal response spectrum method, as defined in EC8. All buildings were designed for the combination of vertical loads 1.35G+1.50Q, as well as the seismic combination G+0.3Q±E, (where G, Q are the dead and live loads, and E is the seismic action expressed by the simultaneous application of the design spectrum of EC8 along the direction of axes x and y). The design of the structural members was made following the provisions of EC2 and EC8, utilizing the professional program for R/C building analysis and design RAF [41].

After the elastic modeling and design of the bare buildings, the three subsets mentioned above (bare, infilled buildings, buildings with pilotis) were created and their nonlinear behavior was simulated, in order to analyze them by means of NTHA. The modeling of the structures' nonlinear behavior was made using lumped plasticity models (plastic hinges at the column and beam ends, as well as at the base of the walls). The Modified Takeda hysteresis rule [42] was adopted in order to model the material inelasticity of the structural members. Moreover, the effects of axial load-biaxial bending moments ($P$-$M_1$-$M_2$) interaction at columns and walls hinges were taken into consideration. The yield moments of the R/C elements and the parameters which were necessary for the determination of the $P$-$M_1$-$M_2$ interaction diagram of the vertical R/C elements' cross sections were computed using the XTRACT software [43].

Regarding the infill walls' modeling, in the present study, the equivalent diagonal strut model was adopted. This model is one of the most well-known and documented in the relevant literature macro-models [44-46]. It does not account for the local failure, but it participates in the building's global collapse mechanism, which is the main objective of the present study. In particular, each infill panel was modeled as single equivalent diagonal strut with stress-strain diagram according to the model proposed by Crisafulli [47] (Fig. 1). Figure 1 illustrates the simulation of the masonry infills based on the Crisafulli model, along with all the basic parameters used to define the properties of the diagonal struts. Note that the values of these parameters were computed with the aid of the code provisions given in EC6 [48].

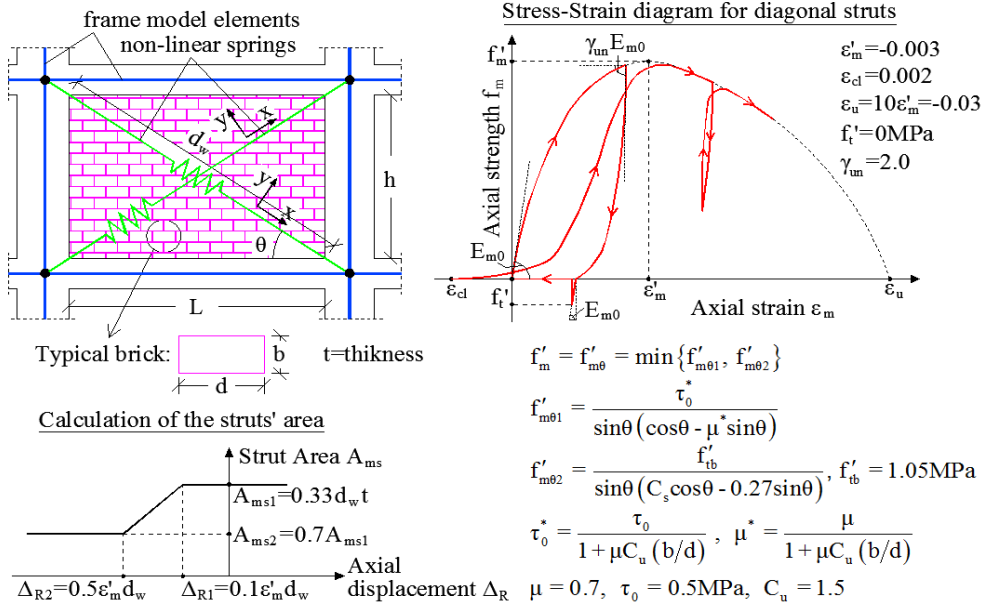

**Fig 1.** Simulation of the masonry infill response using the method of diagonal struts

### 2.3 Inputs Parameters

The Machine Learning methods are computational structures which are capable of approaching the solution of multi-parametric problems. This feature gives the flexibility to select the number of the parameters (input parameters) through which a problem can be formulated. For the present investigation's purposes, both structural and seismic parameters were chosen in order to adequately describe the problem. Considering the structural parameters, four macroscopic characteristic, which are considered crucial for the vulnerability assessment of existing 3D R/C buildings were selected: the total height of buildings $H_{tot}$, the ratios of the base shear that is received by R/C walls (if they exist) along two horizontal orthogonal directions x and y (ratio $n_{vx}$ and ratio $n_{vy}$) and the structural eccentricity $e_0$ (Table 1). As regards the seismic parameters, it must be noticed that there are many definitions of them, which are obtained from the accelerograms records. For the present study, the 14 seismic parameters presented in Table 2 have been chosen (e.g. [49-50]), in an attempt to select the ones widely used by the relevant literature to describe better the seismic excitations and their impact to structures.

**Table 2.** The selected seismic (ground motion) parameters and the ranges of their values corresponding to the 65 earthquakes

| Ground Motion Parameter | Minimum Value | Maximum Value |
|---|---|---|
| **P**eak **G**round **A**cceleration - **PGA** | 0.004g | 0.822g |
| **P**eak **G**round **V**elocity - **PGV** | 0.86 cm/sec | 99.35 cm/sec |
| **P**eak **G**round **D**isplacement - **PGD** | 0.36 cm | 60.19 cm |
| Arias Intensity $I_a$ | ≈0.0 m/sec | 5.592 m/sec |
| **S**pecific **E**nergy **D**ensity - **SED** | 1.24 cm$^2$/sec | 16762.8 cm$^2$/sec |
| **C**umulative **A**bsolute **V**elocity - **CAV** | 14.67 cm/sec | 2684.1 cm/sec |
| **A**cceleration **S**pectrum **I**ntensity - **ASI** | 0.003 g·sec | 0.633 g·sec |
| **H**ousner **I**ntensity - **HI** | 3.94 cm | 317.6 cm |
| **E**ffective **P**eak **A**cceleration - **EPA** | 0.003g | 0.63g |
| $V_{max}/A_{max}$ (**PGV/PGA**) | 0.036 sec | 0.336 sec |
| **P**redominant **P**eriod - **PP** | 0.077 sec | 1.26 sec |
| **U**niform **D**uration - **UD** | ≈0.0 sec | 17.68 sec |
| **B**racketed **D**uration - **BD** | ≈0.0 sec | 61.87 sec |
| **S**ignificant **D**uration - **SD** | 1.74 sec | 50.98 sec |

## 2.4 Output Parameters - Targets

The exported result of the solution of the problem which is examined in the present paper is the estimation of the seismic damage state of R/C buildings, so a reliable measure that can adequately quantify their damage response must be adopted as a target (output parameter) for the Machine Learning algorithms. More specifically, the 90 buildings presented above were analyzed by means of NTHA for a suite of 65 earthquake ground motions, accounting for the design vertical loads. As a consequence, a total of 5850 NTHA (90 buildings x 65 earthquake records) were conducted in the present research. The analyses were performed using the computer program Ruaumoko [51]. Regarding the selection of the input earthquake motions, each of these consists of a pairs of horizontal bidirectional seismic components, obtained from the PEER [52] and the European strong-Motion database [53]. The selection of the records was made bearing in mind the coverage of a large variety of realistic values for the 14 ground motion parameters considered as inputs. In Table 2 the range of the ground motion parameters' values that correspond to the 65 chosen strong motions is depicted. For the calculation of the above seismic parameters, the computer program SeismoSignal [50] was utilized.

For each one of the nonlinear analyses, the assessment of the seismic damage was determined. In particular, the estimation of the seismic damages that are expected to occur in structural members of R/C buildings is accomplished through the calculation of certain measures which try to quantify the severity of the damage. The choice of a reliable damage measure, that can adequately capture the damage level of the building, is a very difficult task, since it depends on numerous parameters. The present research study, in order to express the buildings' seismic damage, adopts the Maximum Interstory Drift Ratio (MIDR). More specifically, MIDR corresponds to the maximum story's drift among the perimeter frames and it is calculated according to Fig.2. The MIDR, which is extensively used as an effective indicator of structural and nonstructural damage of R/C buildings (e.g. [54-55]), has been adopted by many researchers for the assessment of the structures' inelastic response.

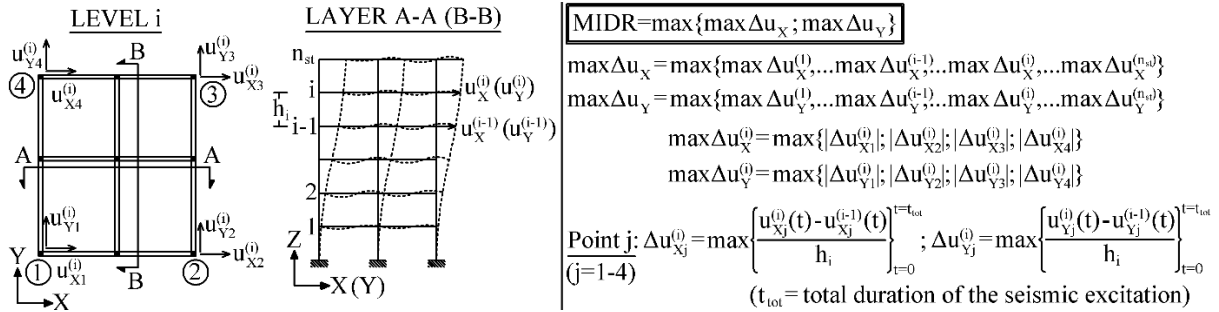

**Fig. 2** Determination of the MIDR in the case of a n-story 3-D building with arbitrary plan-view

## 3. Presentation of used Machine Learning algorithms

In order to identify the most effective algorithm that is capable to predict the R/C buildings' seismic damage with high accuracy, an extensive comparison with the most widely used supervised ML models was made. A comprehensive review of the comparison models is summarized as follows:
1. **Light Gradient Boosting Machine:** is a gradient boosting framework based on decision trees to increases the efficiency of the model and reduces memory usage [56].

2. **Gradient Boosting Regressor:** This method produces an ensemble prediction model by a set of weak decision trees prediction models. It builds the model smoothly, allowing at the same time the optimization of an arbitrarily differentiable loss function [57].
3. **Random Forest Regressor:** A Random Forest is a meta-learner that builds a number of classifying decision trees on various sub-samples of the dataset and uses averaging to improve the predictive accuracy and to control over-fitting [58].
4. **Extra Trees Regressor:** Extra Trees is an information-based learning methodology. Specifically, it is an ensemble machine learning algorithm that combines the predictions from many decision trees [59].
5. **k-Nearest Neighbors Regressor:** k-Nearest Neighbors Regressor is a similarity-based learning algorithm, according to which the target is predicted by local interpolation of the targets associated with the nearest neighbors in the training set [60].
6. **Linear Regression:** Linear Regression is a model that assumes a linear relationship between the input variables (x) and the output variable (y), so that (y) can be calculated from a linear combination of the input variables (x). In linear regression, relationships are modeled using linear prediction functions whose unknown model parameters are estimated from the probability distribution of the prediction values [61].
7. **Bayesian Ridge:** Bayesian Ridge is a type of linear regression algorithm that uses probability distributions rather than point estimates in order to solve a regression problem [62].
8. **Ridge Regression:** Ridge Regression is a regression method that does not provide confidence limits. It uses regularization L2-norm in order to solve a high covariance problem, even if the errors come from an abnormal distribution [63].
9. **Decision Tree Regressor**: A decision tree is a tree-based model including chance event outcomes, resource costs, in order to displays conditional control statements. Each node represents an attribute, each branch represents the outcome of an attribute test, and each leaf represents the decision taken after computing all attributes. The paths from the root to leaf represent the regression process [64].
10. **AdaBoost Regressor:** It is a meta-learner that begins by fitting a regressor on the original dataset and then fits additional copies of the regressor on the same dataset where the weights of instances are adjusted according to the error of the current prediction [65].
11. **Elastic Net:** The Elastic Net is a normalized regression method to fit data that linearly combines the L1 and L2 norms of the lasso and ridge regression methods [66].
12. **Lasso Regression:** Least Absolute Shrinkage and Selection Operator Lasso Regression is a type of linear regression methodology that uses a shrinkage technique in which data are shrunk to a central point, such as the average value [67].
13. **Orthogonal Matching Pursuit:** Orthogonal Matching Pursuit is a sparse approximation algorithm which finds the optimal multidimensional data projection fitting the data with high accuracy [68].
14. **Huber Regressor:** Huber Regressor is a regression method which defines a threshold based on the distance between target and prediction that makes the loss function switch from a squared error to an absolute one [69].
15. **Least Angle Regression:** Least Angle Regression is a linear regression algorithm for fitting high-dimensional data. The solution consists of a curve denoting the solution for each value of the L1 norm of the parameter vector in which the estimated parameters are increased in a direction equiangular to each one's correlations with the residual [70].

## 4. Comparative Assessment of the ML Methods

The abovementioned ML techniques were utilized for the statistical analysis of the training datasets in order to estimate their predictability in the estimation of the buildings' seismic damage. The following regression metrics were used to compare the results and to detect the ML algorithm which is the most efficient:

**Coefficient of Determination - $R^2$.** In order to express the correlation between two random variables, $R^2$ is used which is expressed in terms of percentage. This metric gives the rate of variability of the Y values calculated by X and vice versa. $R^2$ is defined as follows:

$$R^2 = 1 - \frac{\sum_{i=1}^{n}(Y_i - \hat{Y}_i)^2}{\sum_{i=1}^{n}(Y_i - \bar{Y}_i)^2} \tag{1}$$

where $Y_i$ are the observed values of the dependent variable, $\hat{Y}_i$ are the estimated values of the dependent variable, $\bar{Y}$ is the arithmetic mean of the observed values and n is the number of observations. $R^2$ attains values in the interval [0,1], with optimal performance when its values approach the unit, indicating that the regression model adapts optimally to the data.

**Mean Absolute Error – MAE.** MAE is the measure that quantifies the error between the estimated and the observed values. It is calculated by the formula:

$$MAE = \frac{1}{n}\sum_{i=1}^{n}|f_i - y_i| = \frac{1}{n}\sum_{i=1}^{n}|e_i| \tag{2}$$

where $f_i$ is the estimated values and $y_i$ is the observed ones. The average of the absolute value of the difference between these values is defined as the absolute error of their relation $|e_i| = |f_i - y_i|$.

**Mean Square Error – MSE.** MSE is the basic comparison measure that calculates how well a model approaches the number of control examples in a regression process. It is given by the following formula:

$$MSE = \frac{1}{n}\sum_{i=1}^{n}(\hat{Y}_i - Y_i)^2 \tag{3}$$

where $Y$ is an observed value and $\hat{Y}$ is an estimated value for the *n* predictions.

**Root Mean Squared Error – RMSE.** RMSE calculates the average error of the predicted values in relation to the actual values. RMSE is based on the following formula:

$$RMSE = \sqrt{\frac{1}{n}\sum_{j=1}^{n}(P_{(ij)} - T_j)^2} \tag{4}$$

where $P_{(ij)}$ is the value predicted by program i for a simple hypothesis j and $T_j$ is the target value for the simple hypothesis j. The success of a regression model requires extremely small values for the RMSE, while the best case (absolute correlation between actual and predicted values and therefore absolute success of the model) is achieved when $P_{(ij)} - T_j = 0$.

**Mean Absolute Percentage Error – MAPE.** MAPE provides an objective measure of the estimation error as a percentage of demand (e.g. the estimation error is on average 10% of actual demand) without depending on the order of magnitude of demand. It is given by the following formula where $A_t$ is the actual value and $F_t$ is the forecast value:

$$MAPE = 100 \sum_{t=1}^{T} \frac{\left[\frac{|A_t - F_t|}{A_t}\right]}{T} \tag{5}$$

Generally speaking, RMSE gives more importance to the highest errors, hence it is more sensitive to outliers, whereas, on the other hand, MAE is more robust to outliers. RMSE and MSE work on the principle of averaging the errors, while MAE's calculation is based on the median of the error. Finally, MAPE is a very intuitive interpretation in terms of relative error.

In order to confirm the effectiveness of the ML algorithms, extensive ML tests were performed and the comparative results (ranked form the most efficient to the least efficient method) obtained for each one of the three datasets in terms of the abovementioned metrics are presented in the following Tables 4, 5 and 6:

**Table 4.** Performance metrics of the compared algorithms for the bare buildings (dataset ROW_FORM_BARE)

| Machine Learning Algorithm \ Regression Metric | $R^2$ | MAE | MSE | RMSE | MAPE | TT (Sec) |
|---|---|---|---|---|---|---|
| **Light Gradient Boosting Machine** | **0.9076** | **0.1722** | **0.0867** | **0.2902** | **0.1899** | **0.082** |
| Gradient Boosting Regressor | 0.8968 | 0.1904 | 0.0968 | 0.3068 | 0.2452 | 0.205 |
| Random Forest Regressor | 0.8883 | 0.1887 | 0.1035 | 0.3184 | 0.1752 | 0.823 |
| Extra Trees Regressor | 0.8840 | 0.1884 | 0.1064 | 0.3237 | 0.1706 | 0.657 |
| k-Nearest Neighbors Regressor | 0.8343 | 0.2406 | 0.1542 | 0.3875 | 0.2377 | 0.065 |
| Linear Regression | 0.8312 | 0.2757 | 0.1585 | 0.3939 | 0.5849 | 0.019 |
| Bayesian Ridge | 0.8312 | 0.2757 | 0.1588 | 0.3941 | 0.5835 | 0.018 |
| Ridge Regression | 0.8283 | 0.2768 | 0.1622 | 0.3981 | 0.5804 | 0.017 |
| Decision Tree Regressor | 0.7897 | 0.2565 | 0.1941 | 0.4378 | 0.2318 | 0.024 |
| AdaBoost Regressor | 0.7721 | 0.3527 | 0.2099 | 0.4578 | 0.9696 | 0.143 |
| Elastic Net | 0.7650 | 0.3100 | 0.2224 | 0.4675 | 0.3449 | 0.020 |
| Lasso Regression | 0.7647 | 0.3100 | 0.2227 | 0.4678 | 0.3484 | 0.018 |
| Orthogonal Matching Pursuit | 0.7550 | 0.3202 | 0.2318 | 0.4776 | 0.3477 | 0.018 |
| Huber Regressor | 0.7378 | 0.3438 | 0.2478 | 0.4919 | 0.7261 | 0.065 |
| Least Angle Regression | 0.5082 | 0.4422 | 0.4716 | 0.5778 | 1.6721 | 0.021 |

*\* TT (Sec)=Training Time in seconds*

**Table 5.** Performance metrics of the compared algorithms for the infilled buildings (dataset ROW_FORM_FULL-MASONRY)

| Regression Metric<br>Machine Learning Algorithm | $R^2$ | MAE | MSE | RMSE | MAPE | TT (Sec) |
|---|---|---|---|---|---|---|
| **Light Gradient Boosting Machine** | **0.7833** | **0.1535** | **0.2979** | **0.3861** | **0.9794** | **0.078** |
| Bayesian Ridge | 0.7385 | 0.2045 | 0.3217 | 0.4217 | 1.5333 | 0.016 |
| Ridge Regression | 0.7367 | 0.2049 | 0.3230 | 0.4228 | 1.4272 | 0.017 |
| Linear Regression | 0.7366 | 0.2052 | 0.3240 | 0.4230 | 1.4559 | 0.017 |
| Least Angle Regression | 0.7342 | 0.2082 | 0.3249 | 0.4247 | 1.5213 | 0.019 |
| k-Nearest Neighbors Regressor | 0.6760 | 0.1549 | 0.3447 | 0.4532 | 0.2621 | 0.063 |
| Elastic Net | 0.6423 | 0.2640 | 0.3752 | 0.4868 | 2.3327 | 0.017 |
| Orthogonal Matching Pursuit | 0.6378 | 0.2504 | 0.3786 | 0.4880 | 1.4332 | 0.016 |
| Decision Tree Regressor | 0.6352 | 0.2614 | 0.3809 | 0.4893 | 1.4516 | 0.017 |
| Lasso Regression | 0.6300 | 0.2711 | 0.3814 | 0.4941 | 2.2259 | 0.018 |
| Huber Regressor | 0.5998 | 0.2856 | 0.4016 | 0.5117 | 2.4230 | 0.060 |
| Gradient Boosting Regressor | 0.4942 | 0.1759 | 0.4125 | 0.5323 | 0.5681 | 0.182 |
| Random Forest Regressor | 0.3898 | 0.1538 | 0.4435 | 0.5410 | 0.2347 | 0.764 |
| Extra Trees Regressor | 0.2019 | 0.1567 | 0.5402 | 0.5937 | 0.2384 | 0.633 |
| AdaBoost Regressor | 0.0737 | 0.3369 | 0.6196 | 0.6696 | 4.8885 | 0.117 |

*TT (Sec)=Training Time in seconds*

**Table 6.** Performance metrics of the compared algorithms for the buildings with pilotis (dataset ROW_FORM_PILOTIS)

| Regression Metric<br>Machine Learning Algorithm | $R^2$ | MAE | MSE | RMSE | MAPE | TT (Sec) |
|---|---|---|---|---|---|---|
| **Light Gradient Boosting Machine** | **0.8943** | **0.2450** | **0.1999** | **0.4410** | **0.2660** | **0.080** |
| Extra Trees Regressor | 0.8812 | 0.2634 | 0.2251 | 0.4698 | 0.2304 | 0.660 |
| Random Forest Regressor | 0.8792 | 0.2688 | 0.2274 | 0.4726 | 0.2295 | 0.808 |
| Gradient Boosting Regressor | 0.8717 | 0.2884 | 0.2405 | 0.4854 | 0.3973 | 0.197 |
| Decision Tree Regressor | 0.8012 | 0.3373 | 0.3768 | 0.6093 | 0.2777 | 0.027 |
| Linear Regression | 0.7607 | 0.4667 | 0.4580 | 0.6704 | 1.6584 | 0.021 |
| Ridge Regression | 0.7603 | 0.4653 | 0.4604 | 0.6717 | 1.6361 | 0.015 |

| Regression Metric<br>Machine Learning Algorithm | R² | MAE | MSE | RMSE | MAPE | TT (Sec) |
|---|---|---|---|---|---|---|
| Bayesian Ridge | 0.7602 | 0.4655 | 0.4603 | 0.6717 | 1.6418 | 0.019 |
| Least Angle Regression | 0.7423 | 0.4920 | 0.4876 | 0.6925 | 1.6390 | 0.021 |
| AdaBoost Regressor | 0.7411 | 0.5538 | 0.4823 | 0.6919 | 2.5560 | 0.145 |
| k-Nearest Neighbors Regressor | 0.7350 | 0.3926 | 0.5102 | 0.7037 | 0.4030 | 0.063 |
| Huber Regressor | 0.6656 | 0.5199 | 0.6555 | 0.7974 | 1.3578 | 0.063 |
| Elastic Net | 0.6638 | 0.5223 | 0.6512 | 0.7975 | 1.0582 | 0.016 |
| Lasso Regression | 0.6520 | 0.5394 | 0.6736 | 0.8112 | 1.1689 | 0.017 |
| Orthogonal Matching Pursuit | 0.6402 | 0.5574 | 0.6961 | 0.8255 | 1.2555 | 0.017 |

*\* TT (Sec)=Training Time in seconds*

Tables 4, 5, and 6 clearly show the superiority of the Light Gradient Boosting Machine (LightGBM) algorithm, which excels in all metrics, while the performance error remains very low compared to the other approaches. Specifically, the accuracy of the LightGBM, exceeds on average the second-best method by almost 3.5%, while the recorded error is significantly smaller. These features are clearly demonstrated by the very high-performance results that it has achieved, as well as its ability to generalize to new unknown situations and to effectively model real-world data. Specifically, the results revealed that using LightGBM it is possible to correlate sophisticated parameters in a simple way and to solve dynamic problems like the prediction of the R/C buildings' seismic response with high accuracy and with an affordable computational cost.

## 5. Description of the implementation of the most efficient algorithm: LightGBM

In the following, a thorough description, along with analytical details, of the implementation of the most efficient ML algorithm (LightGBM) are given. LightGBM [71] is an information-based learning methodology, which belongs to the class of gradient boosting algorithms and uses a learning algorithm based on regression trees. Regression trees are a simple, easy-to-interpret technique that works best in single-dimensional data analysis (not multidimensional data such as photos, videos, etc.). Considering a set of the form $(x_i, y_i)$ for $i = 1, 2, \ldots, N$ with $x_i = (x_{i1}, x_{i2}, \ldots, x_{ip})$ and for $j = 1, 2, \ldots, p$, the construction of a regression tree is defined as follows:
1. The set of target variable's values $y_i$ is divided into $M$ regions $R_1, R_2, \ldots, R_M$
2. The variable is modeled as a constant $c_m$ in each region so that:

$$f(x) = \sum_{m=1}^{M} c_m I(x \in R_m) \quad (6)$$

Having as a criterion of minimization the sum of the squares $\sum(y_i - f(x_i))^2$ it is easy to calculate the optimal $\hat{c}_m$, which is the average of $y_i$ in the region $m$:

$$c_m = ave(x_i|y_i \in R_m) \tag{7}$$

The problem which arises is that using the sum of the squares in order to find the best results, the algorithm becomes extremely time-consuming. For this reason, another approach is usually used, according to which in each step the target variable is divided into two areas through two branches, a variable $X_j$ and the separation point s are selected, which results in the largest reduction in the sum of squares. Essentially, in this way a variable $j$ and a point $s$ are sought, in order to minimize the following function:

$$\sum_{x_i \in R_1(j,s)} (y_i - c_1)^2 + \sum_{x_i \in R_2(j,s)} (y_i - c_2)^2 \tag{8}$$

where $R_1(j,s) = \{X|X\_j \leq s\}$ και $R_2(j,s) = \{X|X\_j > s\}$. Then, the process is repeated for each area created. The question that arises is how big the trees should be. Note that a large tree will be very specialized in data resulting in a low predictive ability for new data that they have never seen before, while a small tree may not have been properly trained resulting in yielding unsatisfactory results. One solution to the problem is to set a minimum threshold and only if the reduction in the sum of squares achieved by the division is larger than the threshold the separation takes place. This strategy is not always optimal, as a bad initial separation can then lead to a very good next one. The strategy that works best is pruning the tree. The idea is to grow a tree with a predetermined number of nodes and then to prune it using a criterion based on the complexity of the tree as follows:

1. Firstly, a tree is trained at least $T \subset T_0$, which can be any tree that resulted from the pruning of the tree $T_0$.
2. Setting the terminal nodes of $T$, with the node $m$ representing the region $R_m$, then:

$$\hat{c}_m = \frac{1}{N_m} \sum_{x_i \in R_m} y_i \tag{9}$$

$$Q_m(T) = \sum_{x_i \in R_m} (y_i - \hat{c}_m)^2 \tag{10}$$

$$Ca(T) = \sum_{m=1}^{|T|} Q_m(T) + a|T| \tag{11}$$

Essentially, the first term of the function $Ca$ measures how well the tree adapts to the training data (small values indicate good adaptation) and the second term measures the complexity of the tree. The parameter $\alpha \geq 0$ indicates the counterpoint between complexity and good fit of the tree. For $\alpha = 0$ the resulting tree is $T_0$, as no cost is added for each node included in the tree. As the parameter $\alpha$ grows, the cost of the tree complexity increases, so it results in smaller trees which do not adapt as well to the training data. The smaller the parameter $\alpha$ the larger the tree that is constructed, resulting often in overfitting in the training data and, consequently, in a poor performance for other data sets.

As mentioned above, LightGBM is a gradient boosting algorithm. The Boosting technique is based on the creation of successive trees. Each tree is trained using information from previous trees. The algorithm works as follows:

1. For each observation in the set of training data $f(x) = 0$ and $\varepsilon_i = y_i$ is set.
2. In each round $k$ a tree $\widehat{f^k}$ with $d$ nodes is trained, having as a response variable the residuals of the operation (what is left over from the previous regression round) which are denoted by $\varepsilon_i$.
3. A pruned version of the new tree is added so that:

$$\hat{f}(x) \leftarrow \hat{f}(x) + \lambda \widehat{f^k}(x) \tag{12}$$

4. Respectively:

$$\varepsilon_i \leftarrow \varepsilon_i - \lambda \widehat{f^k}(x) \tag{13}$$

5. Repeating the process from step 2 for $K$ times ($K$ is defined by the user) the final form of the model is obtained:

$$\hat{f}(x) = \lambda \sum_{k=1}^{K} \widehat{f^k}(x) \tag{14}$$

In order the Boosting technique to be effective, the user must specify the number of trees to be created, the parameter $\lambda$ and the number of nodes in each tree. A large number of trees can easily be over-adapted to training data resulting in a poor generalization ability. The $\lambda$ parameter determines how fast the model will learn. Typical values of $\lambda$ are from 0.001 to 0.1. The number of nodes controls the complexity of each tree. Often, trees of a single division, also known as branches, are satisfactory because the learning in the model is done slowly and in a controlled way.

The Gradient Boosting technique is an extension of the Boosting technique, combining two methods, the Gradient Descent algorithm and the Boosting technique. Gradient Descent is a first-class optimization method. In order to find the total minimum of a function using this technique, its derivative is firstly calculated and then the inverse process of finding the derivative is used. The derivative measures how much the value of a function $J(\theta)$ will change if the variable $\theta$ changes slightly. It is essentially the slope of the function. High values of the function indicate a large slope and therefore a large change in the value of $(\theta)$ for small changes of $\theta$. This algorithm is iterative, namely it initializes a random value in $\theta$, calculates the derivative of the function at the given point and modifies $\theta$ so that:

$$\theta = \theta - \rho \frac{dj}{d\theta} \tag{15}$$

where the parameter $\rho$ determines how fast it will move in the negative direction of the derivative. The process is repeated until the algorithm converges.

In the case of Gradient Boosting, the algorithm suggests training trees in the negative derivative of the loss function. For example, taking as a loss function the sum of the squares of the residuals $\varepsilon_i$ divided by 2 so that:

$$L(y_i, \hat{y}_i) = \frac{1}{2} \sum_{i=1}^{N} (y_i - \hat{y}_i)^2 \qquad (16)$$

Calculating the derivative:

$$\frac{dL(y_i, \hat{y}_i)}{d\hat{y}_i} = \hat{y}_i - y_i \qquad (17)$$

That is, the negative derivative of the loss function equals to the residuals $\varepsilon_i$. So, essentially, the process involves training a tree based on the $\varepsilon_i$ residuals, to which a pruned by $\rho$ version of the new tree is added. In this way, the Gradient Boosting technique adds successive trees at any given time $t$ to the negative derivative of the loss function so that:

$$\hat{y}_i^{(t)} = \sum_{t=1}^{K} f_t(x_i), \; f_t \in F \qquad (18)$$

where $F = \{f(x) = w_{q(x)}\}$ and $q: R^m \to T, w \in R^T$ that q represents the structure of each tree, $T$ represents the number of leaves and each $f_t$ corresponds to an independent tree structure $q$ with the leaf weights being denoted as $w$. In the LightGBM technique, trees of different structure $q$ are combined, with the structure of each tree being the number of nodes that are created. The loss function that is minimized at any time $t$ is given by the formula:

$$L^{(t)} = \sum_{i=1}^{n} l(y_i, \hat{y}_i^t) + \sum_{k=1}^{T} \Omega_{f(t)} \qquad (19)$$

The first term measures how well the model adapts to the training data (small values indicate good adaptation) and the second term measures the complexity of each tree, where a new term is introduced in addition to the number of leaves ($T$), something that results in a reduction in the weights of leaves:

$$\Omega_{f(t)} = \gamma T + \frac{1}{2} \lambda \sum_{j=1}^{T} w_j^2 \qquad (20)$$

The parameter $\gamma$ indicates the penalty value for the growth of the tree, so that large values of $\gamma$ will lead to small trees and small values of $\gamma$ will lead to large trees. The parameter $\lambda$ regulates how well the tree weights will shrink, namely an increase of its value leads to the tree weights' shrinkage. Thus:

$$\hat{y}_i^{(t)} = \sum_{t=1}^{K} f_t(x_i) = \hat{y}_i^{(t-1)} + f_t(x_i) \qquad (21)$$

So, the problem is deciding which $f_t(x_i)$ minimizes the loss function at time $t$:

$$L^{(t)} = \sum_{i=1}^{n} l(y_i, \hat{y}_i^{(t)}) + \sum_{k=1}^{T} \Omega_{f(t)} = \sum_{i=1}^{n} l\left(y_i, \hat{y}_i^{(t-1)} + f_t(x_i)\right) + \sum_{k=1}^{T} \Omega_{f(t)} \qquad (22)$$

From the power series expansion Taylor it follows:

$$f(x + \Delta x) \cong f(x) + f'(x)\Delta x + \frac{1}{2}f''(x)(\Delta x)^2 \tag{23}$$

So the resulting relation is:

$$L^{(t)} \cong \sum_{i=1}^{n} \left[ l(y_i, \hat{y}_i^{(t-1)}) + g_i f_t(x_i) + \frac{1}{2} h_i f_t^2(x_i) \right] + \Omega_{f(t)} \tag{24}$$

where $g_i = d_{\hat{y}_i^{(t-1)}} l(y_i, \hat{y}_i^{(t-1)})$ and $h_i = d^2_{\hat{y}_i^{(t-1)}} l(y_i, \hat{y}_i^{(t-1)})$.

Subtracting the constants, the loss function becomes:

$$L'^{(t)} \cong \sum_{i=1}^{n} \left[ g_i f_t(x_i) + \frac{1}{2} h_i f_t^2(x_i) \right] + \Omega_{f(t)} \tag{25}$$

Putting $I_j = \{i | q(x_i) = j\}$ the set of observations on sheet $j$, the above relation is reformulated as follows:

$$L'^{(t)} \cong \sum_{i=1}^{n} \left[ g_i w_q(x_i) + \frac{1}{2} h_i w_q^2(x_i) \right] + \Omega_{f(t)} = \sum_{i=1}^{T} \left[ \left( \sum_{i \in I_j} g_i \right) w_j + \frac{1}{2} \left( \sum_{i \in I_j} h_i + \lambda \right) w_j^2 \right] + \gamma T \tag{26}$$

Setting $G_j = \sum_{i \in I_j} g_i$ and $H_j = \sum_{i \in I_j} h_i$ the following relation emerges:

$$L'^{(t)} = \sum_{i=1}^{T} \left[ G_j w_j + \frac{1}{2}(H_j + \lambda) w_j^2 \right] + \gamma T \tag{27}$$

Assuming that the structure of the tree $(q(x))$ is known, the optimal weight on each leaf is obtained by minimizing the above relation with respect to $w_j$, so that:

$$w_j = -\frac{G_j}{H_j + \lambda} \tag{28}$$

Subsequently, by replacing $w_j$, the following equation results, which also calculates the quality of the structure of the new tree:

$$L'^{(t)} = -\frac{1}{2} \sum_{j=1}^{T} \frac{G_j^2}{H_j + \lambda} + \gamma T \tag{29}$$

Finally, the algorithm creates divisions using the following function:

$$Gain = \frac{1}{2} \left[ \frac{G_L^2}{H_L + \lambda} + \frac{G_R^2}{H_R + \lambda} - \frac{(G_L + G_R)^2}{H_L + H_R + \lambda} \right] - \gamma \tag{30}$$

where the first fraction is the score of the left part of the separation, the second fraction is the score of the right part of the separation, the third fraction is the score in case that the separation does not take place and $\gamma$ measures the cost of the complexity of the separation.

The process of solving a problem begins with creating a tree and growing it up to a specific user-defined depth. The tree is pruned in the divisions with a negative Gain and, then, a truncated version of the new tree is added to the model. The procedure is repeated for $K$ times ($K$: parameter defined by the user). It is important to note that the LighGBM algorithm, which is characterized by its efficiency, accuracy and speed, creates histograms and uses the generated classes instead of the entire range of each variable's values, achieving a significant reduction in training time. It also grows vertically, which means that it grows at the level of leaf (leaf-wise method, Fig. 3), while other algorithms grow at depth (depth-wise method (Fig. 4)), choosing to grow the leaf with the maximum difference of the cost function. During the leaf-wise tree growth, the algorithm becomes very efficient, as it can significantly reduce the losses, thus gaining accuracy, while at the same time the regression processes are completed quickly.

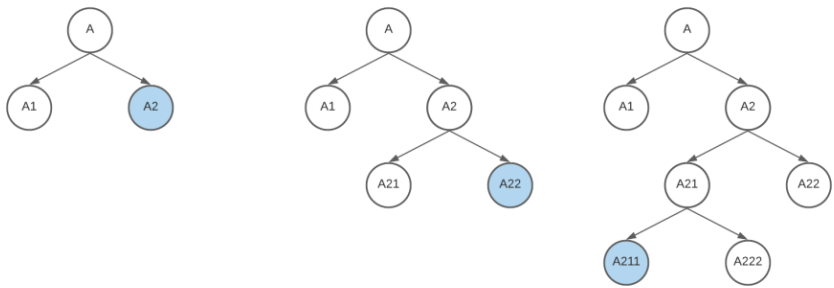

**Fig 3.** Leaf-wise method

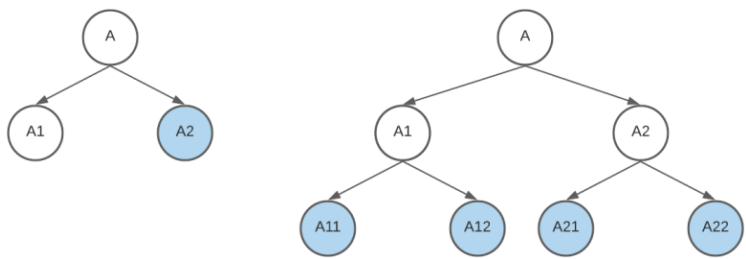

**Fig 4.** Depth-wise method

Another important feature that makes LightGBM one of the most complete and widespread algorithms in Machine Learning is that it does not use all the training data, but a sample of them, which results from the Gradient One Side Sampling method (GOSS). The basic idea of the GOSS methodology focuses on the fact that not all observations contribute the same to the training of the algorithm, since those with a small cost function's first derivative are better trained than those with a large one. Ignoring the observations with a small derivative result in the creation of biased samples and in a definite change in the distribution of data, something which leads to a separation that is greater than the optimal one and to an obvious over-adaptation of the model to the sample. To address the problem, random observations with a small cost function's derivative are selected, which are sorted according to the absolute value of their derivative. Finally, the $a \times 100\%$ with the largest derivative and the $b \times 100\%$ from the rest are selected. For the calculation of the loss function the observations with a small derivative are multiplied by $\frac{1-a}{b}$, thus giving more importance to the poorly trained, without significantly differentiating the distribution of the data. By training only one sample in each iteration, a significant increase in the process of the algorithm learning is achieved, resulting in its fast convergence to the optimal solution. Specifically, for a training set of $T$ with $n$ cases such that $T = \{x_1, x_2, ..., x_n\}$, where each $x_i$ is a vector with dimension $s$ in the space $X_s$. In each iteration of the gradient boosting algorithm, the negative slopes of the cost function in relation to the output of the model are denoted as $G = \{g_1, g_2, ..., g_n\}$. Implementing the GOSS method, the cases are classified according to the absolute values of their degrees in descending order. Thus, a set $A$ with the $a \times 100\%$ larger slopes, a set $A^c$ consisting of $(1 - \alpha) \times 100\%$ cases with the smallest slopes and a subset $B$ with size $b \times |A^c|$ are created. Then, all cases are classified according to the estimated variance cost in vector $V_j(d)$ on the set $A \subset B$, so that:

$$\bar{V}_j(d) = \frac{1}{n}\left(\frac{\left(\Sigma_{x_i \in A_l} g_i + \frac{1-a}{b}\Sigma_{x_i \in B_l} g_i\right)^2}{n_l^j(d)} + \frac{\left(\Sigma_{x_i \in A_r} g_i + \frac{1-a}{b}\Sigma_{x_i \in B_r} g_i\right)^2}{n_r^j(d)}\right) \quad (31)$$

where $A_l = \{x_i \subset A : x_{ij} \subset d\}$, $A_r = \{x_i \subset A : x_{ij} \subset d\}$, $B_l = \{x_i \subset B : x_{ij} \subset d\}$ and $B_r = \{x_i \subset B : x_{ij} \subset d\}$, while the coefficient $\frac{1-a}{b}$ is used to normalize the sum of the slopes above $B$ with respect to the magnitude of $A^c$.

The performance metrics of the LighGBM algorithm for the three datasets considered herein were given in Tables 4, 5, and 6. Generally, the LighGBM algorithm achieves the highest coefficient of determination, while the error fluctuation remains very low in comparison to the other methods. This gives a clear explanation that a large percentage of data points (91% in the first dataset, 78% in the second dataset, and 89% in the third dataset) fall within the results of the regression equation, therefore the method adapts optimally to the data. Note that in the above Tables some of the most valid error metrics are compared, since, in the forecasting procedure by ML methods, the error measurement between the estimated value and the actual value is useful both to assess the performance of the model and to define the objective function of the model. In any case, the LightGBM approach produces the lowest error, which is explained as high overall performance, training stability, and generalization ability. Finally, the algorithm has satisfactory training times, which can be further improved if the training data are pre-sorted.

Diagrams of the methodology, that show its superiority and the way the LightGBM algorithm works, as well as the way of modeling the problem, are presented in the following. The plots of LightGBM algorithm for the dataset of the bare buildings are presented in the following Figs 5-8:

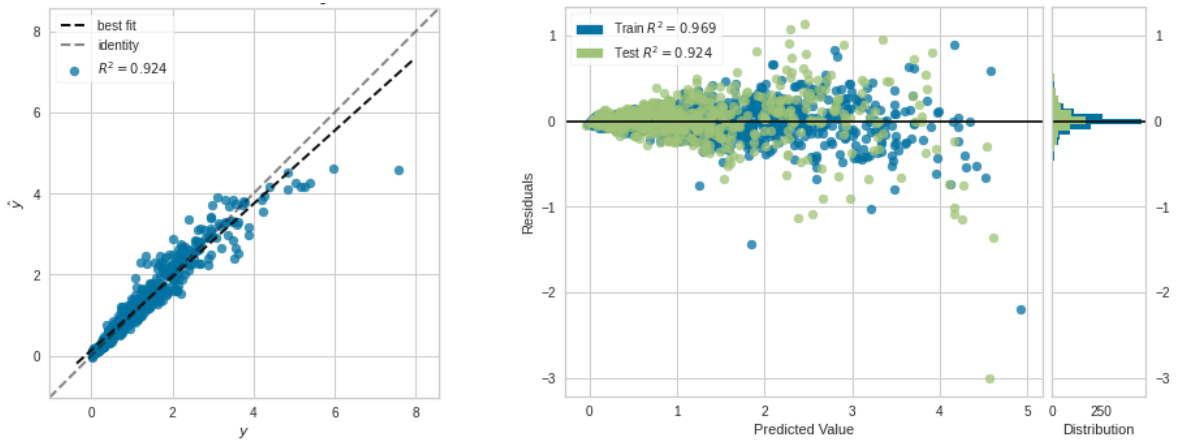

**Fig 5.** Prediction Error for LightGBM for 30 iterations

**Fig 6.** Residuals for LightGBM for 30 iterations

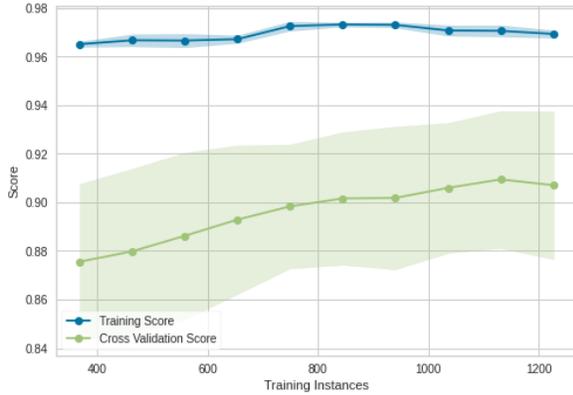
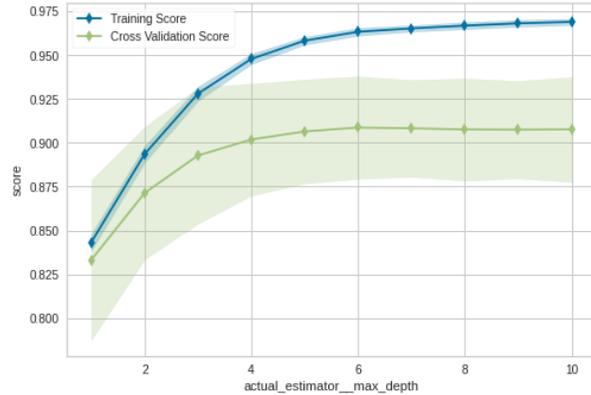

**Fig 7.** Learning curve for LightGBM

**Fig 8.** Validation curve for LightGBM

The plots of LightGBM algorithm for the dataset of the infilled buildings are presented in the following Figs 9-12:

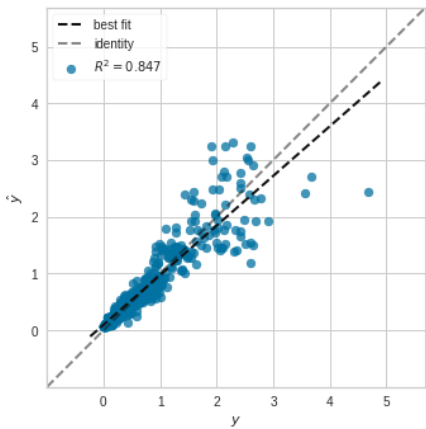
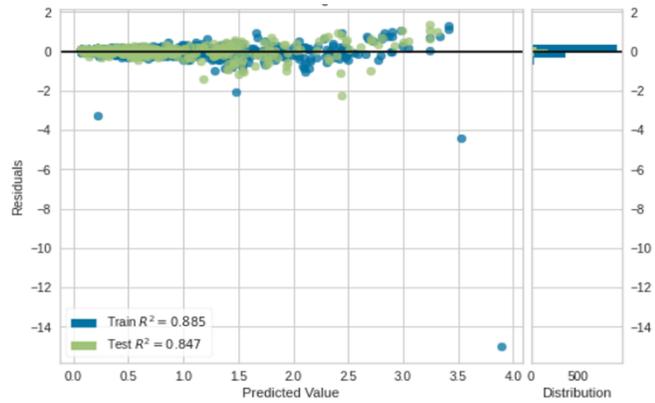

**Fig 9.** Prediction Error for LightGBM for 30 iterations

**Fig 10.** Residuals for LightGBM for 30 iterations

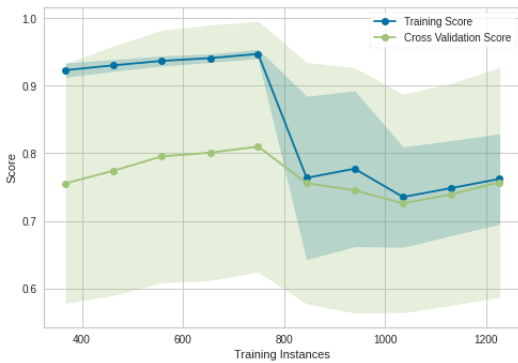
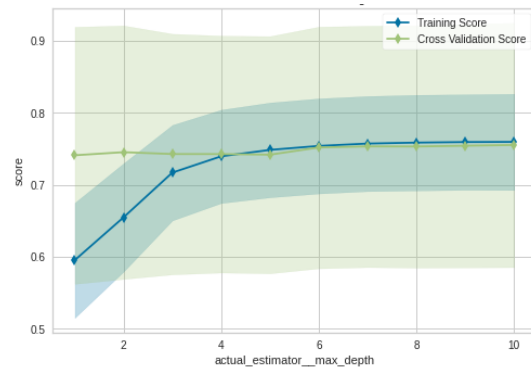

**Fig 11.** Learning curve for LightGBM

**Fig 12.** Validation curve for LightGBM

The plots of LightGBM algorithm for the dataset of the buildings with pilots are presented in the following Figs 13-16:

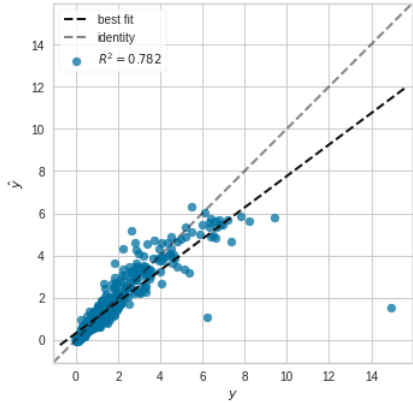 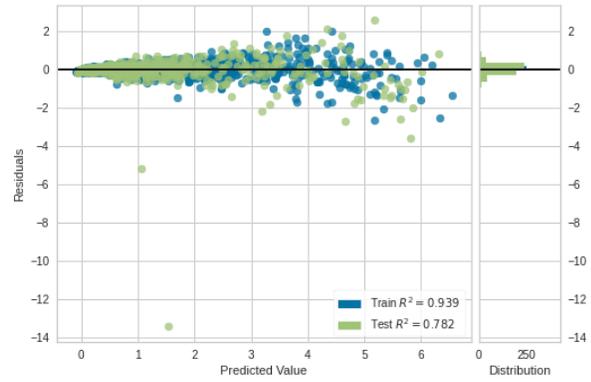

**Fig 13.** Prediction Error for LightGBM for 30 iterations

**Fig 14.** Residuals for LightGBM for 30 iterations

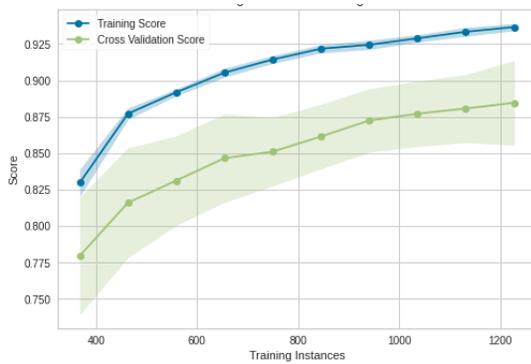 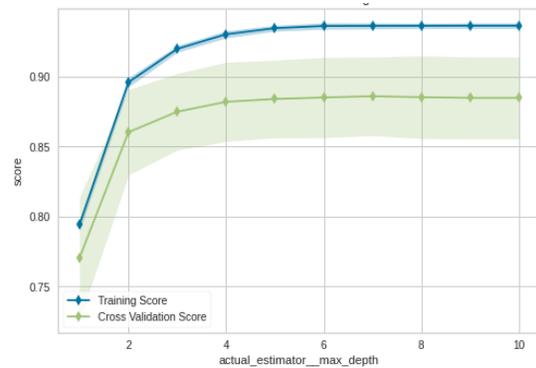

**Fig 15.** Learning curve for LightGBM

**Fig 16.** Validation curve for LightGBM

More specifically, the prediction error plot shows the actual targets from each dataset against the predicted values generated by the model. This allows identifying how much variance exists in the model by comparing them against the 45º line, where the prediction matches exactly the model. Also, the residual plot is a graph that shows the residuals on the vertical axis and the independent variable on the horizontal axis. If the points in a residual plot are randomly dispersed around the horizontal axis, a linear regression model is appropriate for the data; otherwise, a nonlinear model is more appropriate. Moreover, a learning curve is a plot that shows time or experience on the x-axis and learning or improvement on the y-axis. The model is evaluated on the training dataset after each update during training and depicts the measured performance. Finally, the validation curve is a graphical technique that can be used to measure the influence of a single hyperparameter. By looking at this curve, it can be determined if the model is underfitting, overfitting or just-right for some range of hyperparameter values.

## 5. Conclusions

In the present paper an extensive comparative evaluation of a large number of Machine Learning algorithms for the reliable prediction of 3D R/C buildings' seismic response was carried out. In order to accomplish this aim, a large training dataset consisting of 30 R/C buildings with different structural parameters (the number of stories, the structural eccentricity and the ratio of base shear received by R/C walls (if they exist) along the two orthogonal horizontal axes) was selected. The buildings were designed on the basis of provisions of EC8 and EC2. For each one of these buildings three different configurations regarding their masonry infill walls were assumed (without masonry infills, with masonry infills in all stories and with masonry infills in all stories except for the ground story), leading to three different data subsets consisting of 30 buildings each. The selected buildings were analyzed for 65 appropriately chosen real earthquake records using Nonlinear Time History Analyses. As inputs in the process of Machine Learning methods both seismic and structural parameters widely used in the literature were chosen. The well-documented Maximum Interstory Drift Ratio was selected as the damage index for the R/C buildings. Based on the research study's results, the following conclusions can be drawn:

- Historical data can be utilized in order to develop a realistic model, capable to effectively simulate the earthquake response and to predict with great accuracy the seismic damage of structures belonging to different types.
- The general methodology of the proposed procedure uses the most technologically advanced methods in the field of civil engineering and expands them significantly, as it extracts the hidden knowledge found in structural and seismic data in order to add intelligence to the methods of seismic response prediction, as well as to the mechanisms for optimal decision-making related to seismic risk.
- The high generalizability of the LightGBM algorithm, as well as the convergence stability of the proposed methodology, proves that it is capable of performing well even when the problem is multiparametric.
- The GOSS technique used by the LightGBM algorithm handles with great precision the noisy scattered points of incorrect classification, something that other methodologies cannot handle.
- The tree segmentation method utilized by the algorithm leads to results characterized by remarkable prediction, while offering generalization, which is one of the key requirements in the field of machine learning. Moreover, it reduces bias and variance, as well as eliminates overfitting, implementing a robust forecast model.
- The proposed method, as a problem of multiple spatial-temporal variables, argues that machine learning methods can be utilized in order to solve dynamic problems of high complexity with affordable computational costs.
- The proposed procedure constitutes a very promising methodology, which can significantly improve the safety of structures and infrastructure in general under earthquake excitations.

The most important task for the evolution of the proposed methodology is, initially, the process of finding optimization solutions to achieve higher accuracy results. Also, of great importance is the detection of the optimal hyperparameters of the algorithm, in order to enhance the predictive process. Moreover, the training dataset can be expanded to buildings with different structural characteristics and to earthquake records with seismic features of greater range. Finally, the expansion of the methodology with data transformation techniques should be considered, so that the algorithm can locate the optimal representations of the input variables in order to make it easier to extract the useful information.

## 6. References

[1]     ASCE/SEI 41-13. Seismic Evaluation and Retrofit of Existing Buildings. American Society of Civil Engineers (ASCE), Reston, VA; 2014.
[2]     ATC. Earthquake damage evaluation data for California, Redwood City, CA: Applied Technology Council; 1985. ATC-13 Report.